# Great New Design: How Do We Talk about Media Architecture in Social Media

Selena Savic,
Institute of Experimental Design and
Media Cultures (IXDM) FHNW, Basel,
Switzerland
selena.savic@fhnw.ch

## ABSTRACT

In social media, we communicate through pictures, videos, short codes, links, partial phrases. It is a rich, and digitally documented communication channel that relies on a multitude of media and forms. These channels are sorted by algorithms as organizers of discourse, mostly with the goal of channeling attention. In this research, we used Twitter to study the way Media Architecture is discussed within the community of architects, designers, researchers and policy makers. We look at the way they spontaneously share opinions on their engagement with digital infrastructures, networked places and hybrid public spaces. What can we do with all those opinions? We propose here the use of text-mining and machine learning techniques to identify important concepts and patterns in this prolific communication stream. We discuss how such techniques could inform the practice and emergence of future trends.

## KEYWORDS

Machine Learning, Social media, Text-mining, Topic-modelling, Placemaking, Digital infrastructure

## 1 INTRODUCTION

In social media, we communicate through pictures, videos, animated gifs, short codes, links, incomplete sentences, references to other things. It is a rich, and digitally documented communication channel that relies on a multitude of media and forms. Several years ago, architect Mark Kushner observed how social media streams shape the discourse about buildings and render the opinions of the public, the critics and the practitioners closer together[1]. This observation highlights, if not also inflates, the effect of immediacy in social media discourse on popularity of unseen, unusual architectural designs, their consecutive spreading around the world, and adoption by different communities.

Social media communication channels are driven by algorithms as organizers of discourse, mostly with the goal of channelling attention. Continuous and careful redesign of Twitter and Facebook feed algorithms confirm the importance given to the attention of their users. Tweets that the new algorithm ranks as more relevant to the user will appear at the top of the timeline[2]; Facebook news feed is driven by an algorithm that prioritizes posts from friends and family to posts from businesses, brands and media[3].

On the other hand, the sourcing of these discourses programmatically through APIs (application programming interface) and automated account actions has become accessible to anyone with basic programming skills[4]. The sheer quantity of data that can be collected in this way, as well as their short, erratic character escapes the scope of close reading and classical discourse analysis. Data-mining, text-mining and machine learning are increasingly relevant to working with discourse. These approaches inform new, different critical perspectives on questions that have been traditionally addressed in both humanities and engineering. In the context of architecture, it is interesting to imagine how this could inform future practice and future thinking. By looking at how contemporary issues are being discussed, we can get hints at possible future trends and questions that will have to be addressed. We can identify certain social and cultural aspects within the community. We can offer a critical perspective on practice from the angle of how it is discussed.

In this research, we use Twitter posts from a selected group of users to study the way a community gathered around the interest in Media Architecture spontaneously shares opinions on their engagement with digital tools, media and infrastructures. We gathered and text-mined the posts, looking for a network of keywords which would form the basis for a productive discourse. The results of this work are presented in the form of a corpus passport, which document the corpus structure, term-frequency, term co-occurrence and emerging topics. The intention is to facilitate a comparative perspective on different concepts and make explicit the concepts that inform emerging topics.

---

[1] Mark Kushner discusses this and other observations about architecture in his 2014 TED talk, accessible here: https://www.ted.com/talks/marc_kushner_why_the_buildings_of_the_future_will_be_shaped_by_you (accessed on May 3rd 2018)

[2] Twitter blog post about the new timeline algorithm https://blog.twitter.com/official/en_us/a/2016/never-miss-important-tweets-from-people-you-follow.html (accessed on May 4th 2018)

[3] Mar Zuckerber's post announcing the redesign of Facebook news feed https://www.facebook.com/zuck/posts/10104413015393571

[4] A large number of tutorials and guides exist on the Internet, with step-by-step instructions, and relying on open-source tools and and readily available libraries for Python – such as Bonzanini's Mining Twitter Data with Python, available at https://marcobonzanini.com/2015/03/02/mining-twitter-data-with-python-part-1/ (accessed on May 3rd 2018) or The Rickest Ricky's Another Twitter sentiment analysis with Python https://towardsdatascience.com/another-twitter-sentiment-analysis-bb5b01ebad90 (accessed on May 3rd 2018)

We are well aware of the implicit value that case studies of architectural projects have for the advancement of the field of media architecture. At the same time, we observe a certain homogeneity of expressions and topics that leads to a hermeticity of discourse in this community, attuned at supporting social structure of hybrid, networked cities. We therefore felt it is important to engage with the way projects, events, concerns and experiences are discussed in an informal, dynamic way. We look at the way architects, designers, researchers and policy makers spontaneously share opinions on their engagement with digital infrastructures, networked places and hybrid public spaces.

What can we do with all those opinions? How could we identify keywords and patterns that characterize current trends? What could they demonstrate in terms of future trends? To answer these questions, at least in part, we will look into the more general text-mining practice of social media and the kind of patterns and organization researchers are looking for. Which algorithms promise to identify or demonstrate discourse organization, most prominent opinions, and networks of meaning? We then present our methods and our main points of interest. We use machine learning techniques to allow categories and patterns to emerge from data. We discuss in detail our findings and the way they can inform future discourse.

## 2 TEXT-MINING AND ORGANIZING ONLINE DISCOURSE: PRACTICES OF MARKETING, NEWS AND SOCIAL MEDIA ANALYSIS

In this research, we explore the use of text-mining as a data-driven research method that allows categories to emerge from data. This approach can be said to share the goals of certain qualitative methods, such as grounded theory and content analysis. For example, Yu et al. maintain that text mining similarly encourages open-mindedness and discourages preconceptions [18]. Different from grounded theory, our aim is not to create a theory about the way architects talk in social media but rather to experiment with this rich textual material and learn to orient ourselves within it.

Discourse analysis, the field that studies language use (discourse, conversation, communicative event), is typically concerned with keywords, their orientation and assumptions they support. Fairclough focused on discourse analysis of text based on orders of discourse (discourse, genre, style): a network of social practices in its language aspect, rather than grammatical or semantic units [3]. Text is seen as a part of social events with a significant role in the construction of the social values and beliefs. He insisted on keeping a view on text, that which is made explicit, always against the background of that which is left implicit – without expecting to make meaning transparent through analysis.

David M. Berry discussed the computational turn in humanities, in particular literature studies [1]. The digital, he argued, is the new unifying idea in academia and knowledge. With this new idea, reasoning shifts towards a more conceptual or communicative method, a way of thinking that raises different kinds of questions (e.g. how many times a word repeats in a text) and leads to different kinds of findings (e.g. a connection between word collocation and cultural beliefs or values). A similar view is expressed by Jockers, who discussed data mining approaches as *macroanalysis* [6]. He saw its strength in the capacity to zoom in and out (of text, sensor or census data, etc.) and the ability to study information that escapes our attention due to its multitude. Topic modelling or topic detection, a machine learning method to discover topics in text, can be applied to literature on different levels: a corpus, a book, a poem.

Marketing research advocates using social media to engage with customers [14]. As Ohsawa and Yada maintained, the advantage of this approach is in the ease of access to a vast amount of opinions, ideas and interests by diverse people [13] – something that would otherwise require long and detailed preparation of interviews, focus groups and other qualitive approaches. This interest gave rise to a multitude of tools and techniques for sentiment analysis and opinion mining of online textual sources for various recommendation systems, automated personal assistants, customer analysis etc.

A particularly important question when working with social media discourse concerns the way information emerges from textual data. In this respect, machine learning algorithms that facilitate emergence of unlabelled classes and their organisation are of particular interest. For example, self-organizing map (SOM) is a machine learning algorithm introduced by Teuvo Kohonen [7], for ordering high-dimensional statistical data so that alike inputs are mapped closer to each other, illustrating the similarity relationships between different data items (such as text documents) in a familiar and intuitive manner [5]. SOM produces a low-dimensional, discretized representation of the input space, (e.g. topics or words in a corpus) [5].

Researchers have investigated the use of SOM for topic modelling and sentiment analysis. Kohonen himself published extensively on his work with text [5,8,15]. He described how entire documents can be distinguished from each other based on their statistical models [8,9]. In the 1990s, he and his colleagues developed the WEBSOM method to perform a completely automatic and unsupervised full-text analysis of a set of Usenet newsgroup articles [5]. Later, using the SOM on a collection of over one million documents, he demonstrated how his random-projection methods give accurate and fast results in encoding and categorizing documents, without relying on an underlying theoretical model [8].

Lee and Young created word and document cluster maps using SOM to cluster words of similar meaning and documents with overlapping words (similar discourse), on a corpus of news from a Chinese News Site [10]. They developed a search function that retrieves documents based on similarity, a feature that is attuned at explaining impacts of events presented in the news. Using SOM to facilitate analytical inquiries into relationships of words and documents, they demonstrated the power of unsupervised clustering without pre-existent categories.

[5] In machine learning, the process of reducing the number of random variables under consideration by obtaining a set of principal variables, which can be divided on feature selection and feature extraction, is essential for tasks such as topic detection.

Sharma and Dey documented their research in clustering positive and negative online movie reviews using SOM and a supervised learning algorithm, Learning Vector Quantization (LVQ) [16]. They were able to cluster opinions using these techniques with an accuracy of 83% and 89.1% respectively. In another paper co-authored by Dey, Twitter was mined based on bigrams (two-word associations) [17]. Their main contribution is a demonstration of importance of bigram-based analysis for tweet clustering, and thus to topic detection. They use the Hopkins index [6] to demonstrate a measurable increase in clustering tendency and present it using a SOM.

## 3 METHOD: MINING TWITTER DISCOURSE FROM ARCHITECTS, DESIGNERS AND POLICY MAKERS INTERESTED IN MEDIA ARCHITECTURE

For the purpose of this research, we gathered a large collection of tweets from a community of architects, designers, engineers, community managers and policy makers. We used this collection to discover keywords, patterns and topics that emerge from the data and show how these could inform future discussion and practice.

### 3.1. Identifying the Community

We focus on the community that actively shares opinions and experience about media architecture. We identified four main *hashtags* that are specific to the topic, connected to the four major events: Media Architecture Biennale ('#MAB12', '#MAB14', '#MAB16', '#MAB18') and the more general '#mediaarchitecture' hashtag. We used these to search for *tweets* that contain them and identify accounts that use them. Additionally, we include an extended community of accounts that *retweeted* or *liked* tweets that contain those hashtags. Although it is relatively easy to programmatically obtain a list of all accounts that use or interact with a certain hashtag (through Twitter API, with simple scraping tools), we manually reviewed this list of accounts by looking at their profile descriptions[7] and sometimes also their tweets, in order to verify their relevance to the discourse.

In this work, we relied on a technique described by Grandjean, which was used in identification of the Digital Humanities social network [4]. The profiles we selected mentioned, retweeted or liked the hashtags from our list at least once. An equally important criterion was whether the account's profile description confirmed the interest in design and architecture, digital placemaking, smart cities and networked spaces. When this was not sufficient to get an impression, we briefly reviewed the content of the tweets, looking for those that contain opinions or information on media architecture events, projects and techniques. We could observe that the selected profiles mostly belong to architects, designers, engineers, community managers and policy makers, with a mixed presence of practitioners and academics.

Using this approach, we delimited a community of 250 profiles that talk about the experience of digital infrastructures, smart cities, presence of media (facades) in public space, and their effects on urban experience.

### 3.2. Gathering Text and Preparing for Analysis

The analysis presented here extends to all tweets posted from the list of accounts, and not only the tweets that contain the hashtag.

We scraped all available tweets using Twitter API and the Python *tweepy* library, without looking at the content of individual tweets. It is a relatively large corpus of all tweets from users that once or more mentioned the key hashtags or liked other tweets that did so. Our collection contains over 450 000 entries. From those, we only worked with tweets in English (393 104).

Prior to the first, statistical analysis of the text, we performed some simple preprocessing operations. We created three sets of tweets.

In the first set, we preserved special kinds of words like *mentions* (names of Twitter users, beginning with an "@"), *hashtags* (self-declared keywords, beginning with "#") and links. We corrected negations ( "isn't" to "is not") and identified some special groups of characters (e.g. *emoticons*) to be preserved in the text. We converted all text to lowercase, to avoid treating words "Architecture" and "architecture" as different occurrences. We did not apply stemming (rendering words with the same roots into the same entity by removing the ending; e.g. insid-e and insid-er) because we work with the corpus in an open-ended way and stemming could unnecessarily remove some complexities.

In the second set, we did all the previous, and removed mentions, hashtags and links. This corpus thus contains less of the "twitter tone", but is more reliable for inference of meaningful word associations. We also removed punctuation.

We then created a third set of tweets, with stopwords[8] removal based on a general English dictionary enriched with a list of Twitter-specific words (such as 'RE', 'co', 'http', '@', etc).

We addressed all tweets as individual documents in our analysis.

### 3.3. Towards a Corpus Passport: Statistical Analysis of Tweets

The first step in our text-mining approach was to describe the structure of the corpus. For this, word frequency, uniqueness, number of special words as well as word co-occurrence in the corpus are interesting. We used Python *scikit-learn* library

[6] Hopkins index is a way of measuring the cluster tendency of a data set, introduced by Brian Hopkins and John Gordon Skellam in botanical research, in the 1950s
[7] Profile descriptions for Twitter accounts is a short text, 140 characters, that appears under the screen name and the twitter handle
[8] Stopwords are frequently appearing words of a general meaning and orientation (such as 'the', or 'yours') which are filtered in the processing of text.

(CountVectorizer) to extract words and their frequency in the corpus.

*3.3.1. Corpus Structure Analysis – Unique Words, Stop Words.*

We note the number of unique words in the corpus, as well as the number of all words that are not stopwords. This gives a good idea of the specific corpus structure – whether the discourse is rich in unique words or whether it is formed of mainly repeated phrases. Higher level of repetition can be expected to appear in social media through retweets (where people re-post someone else's tweet) and tweet threads (tweets that are reactions, comments and answers to one tweet, where opinions and words tend to repeat a lot).

*3.3.2. Term Frequency and Term Co-occurrence.*

Term frequency is one of the most basic elements of the statistical analysis of a corpus. It gives an idea as to which terms are important in the discourse, but also potentially normalized to such extent that their use does not give a sense of orientation.

Multiple words association offers a different view on the corpus – two and three word associations that appear frequently tell more about the character of the corpus than isolated individual words. We identified most frequent commonly co-occurring words (bigrams and trigrams), using the same Python *scikit-learn* library tools.

## 3.4. Towards a Corpus Passport: Topic Modeling

*3.4.1. Corpus Vectorization*

Further explorations are based on vectorized models of the corpus. This means that the word probability and position is transformed into vector matrices that represent that word's characteristics with numerical values. The most common approaches to corpus vectorization create word2vec and doc2vec models.

Word2vec model is a shallow, two-layer neural network that is trained on linguistic contexts of words [11,12]. It accounts for multiple degrees of similarity words can have in a corpus. Word2vec takes as its input a large corpus of text and produces a vector space with each unique word in the corpus being assigned a corresponding vector in the space. Words that share common contexts in the corpus are located in close proximity to one another in the vector space.

We can extract word similarity from a Word2vec model. Similarity in means that words a closer in meaning, according to a "similarity score" calculated in the vector space. It is interesting to observe this similarity specific to a corpus, and not in general, as small irregularities could imply something specific about the discourse.

Word2vector models can also be used in cluster analysis, and to train neural networks such as SOM. In our work, we tried training a SOM on the model based on words as well as on documents – Doc2vec model. We observed differences in resulting configuration.

*3.4.2. Topic Modelling*

Topic modelling is a form of clustering of words which have a significant relationship in the corpus. The significance of this relationship is determined based on probability distributions of sets of words, each topic characterized by a specific identified distribution. Topic modeling words with individual words, rather than documents. It supports text classification, often used in recommender systems and to uncover themes in texts.

In our work, we tested several different algorithms for identifying topics in the corpus and decided to use Latent Dirichlet Allocation, or LDA model: a generative probabilistic model for collections of discrete data (such as words in corpora of text) [2]. Like some other topic modelling algorithms, LDA automatically indexes, searches, and clusters terms to form unstructured and unlabeled topics – or lists of keywords. In LDA, topics are created using machine learning algorithms to deduce the probability of terms present in each topic and the probability of a topic found in each document through an iterative process. We used Python scikit-learn LatentDirichletAllocation implementation of the algorithm, which enabled us to identify a chosen number of topics (ten) described by a chosen number of keywords (twelve) but also a number prominent documents (ten) to verify the context.

*3.4.3. Self-organizing Maps*

We used the self-organizing map algorithm for data clustering and their graphical representation. It is a computational method for the low-dimensional approximation (i.e. map) of high-dimensional data (i.e. vector space of a word2vec model). Data points with similar features are mapped onto the same region of the map. In our case a data point is a tweet, the high-dimensional input space is the topic space, and the low-dimensional output is a layer of neurons, arranged in a grid to form the map.

We trained two SOM models, one based on words (word2vec model) and the other on documents (doc2vec model) to see how differently they will cluster.

# 4 RESULTS: HOW ARCHITECTS SPEAK ABOUT DIGITAL INFRASTRUCTURES: EMERGING KEYWORDS AND TOPICS

With the interest to identify keywords, topics and patterns discourse in social media, we turned to a community we identified as interested in media architecture. We were looking to establish a relation between corpus structure, occurrence of certain keywords, as well as word co-occurrence and the homogeneity or productivity of discourse. Discourse can be considered productive when terminology is such to enrich and inspire practice, and avoid untheorizable arbitrariness.

In the following text we will present the results of our statistical analysis and topic modelling. We will, when significant, compare these results to properties of other discourses and corpora that we could observe or analyse.

## 4.1. Statistical Analysis Results and Comparison to Other Discourses

*4.1.1. Corpus Passport: Verbosity and Redundancy of Discourse.*

Statistical analysis of corpus structure shows that the corpus contains a reasonable number of unique words – they make 2.51% of the entire corpus (Figure 1).

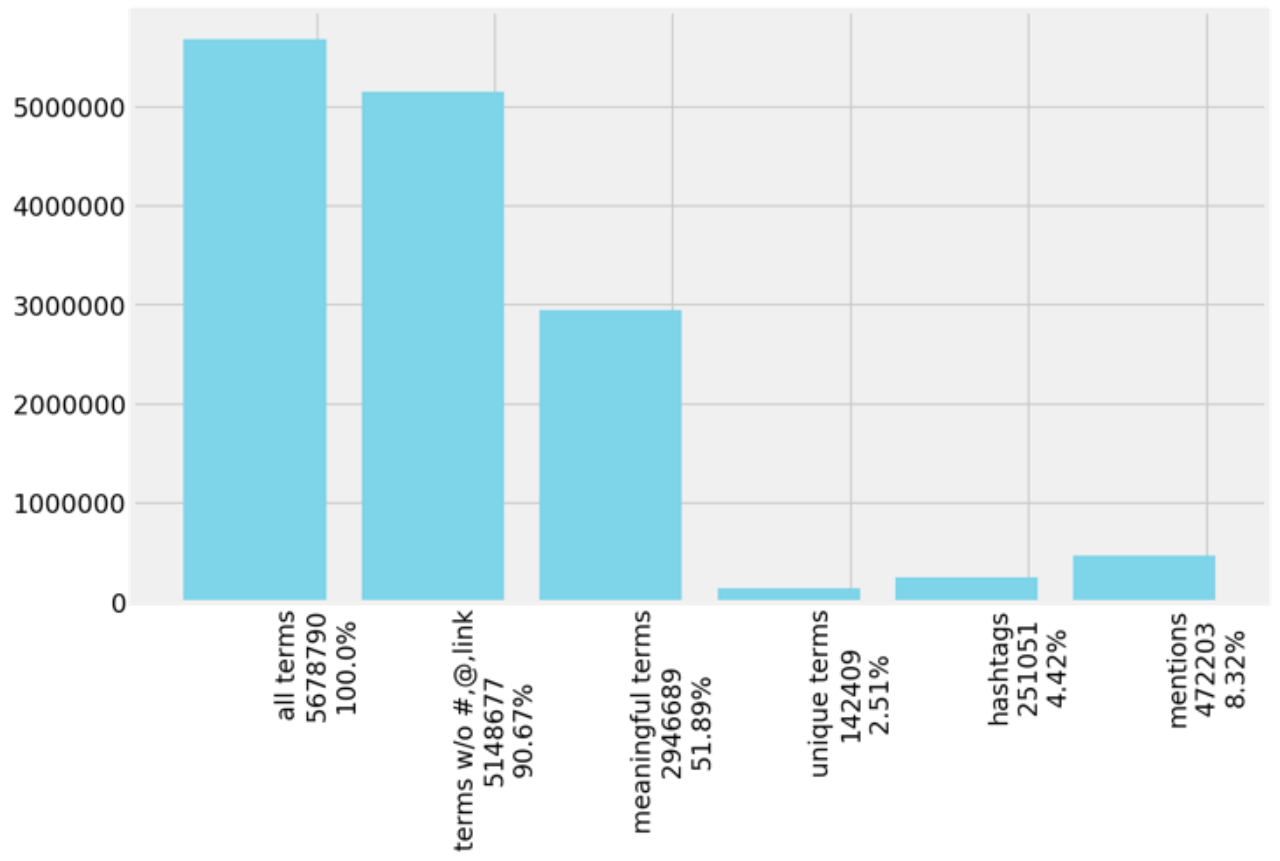

**Figure 1: Visualization of the corpus structure, showing the total number of words, number of unique words and number of all meaningful words**

According to a vocabulary analysis of books available through Project Gutenberg[9], a written novel has anywhere from 2% (The Book of Mormon) to 8,16% (Melville's Moby Dick) and even 17.73% (Lovelace's Lucasta Poems) unique words in a book. For the sake of comparison with more relevant literature, we performed this analysis on a small selection of books from architecture-specific literature that treats computation and digital infrastructures: Keller Easterling's *Extrastatecraft*, Luciana Parisi's *Contagious Architecture* and Benjamin Bratton's *The Stack*. The choice of the books is representative of an approach to computation and architecture through metaphor, taking a topic from the professional domain of architecture – such as urban design or building form – and mapping it to a computational process – execution of space, viral contagion, protocolization of communication. Their vocabulary is made of 11.1% (*Extrastatecraft*), 6.85% (*The Stack*) and 5.96% (*Contagious Architecture*) unique words.

Words specific to Twitter – hashtags and mentions form a relatively large part of the discourse – 12.81% of all words, or almost a quarter (24.54%) compared to the number of meaningful words. Meaningful words are words longer than 3 characters, not belonging to a typical English or Twitter stopword dictionary.

A specific term frequency pattern can be observed in English, but also other languages: the frequency of any word is inversely proportional to its rank in the frequency table. This pattern is known as the Zipf's Law [19]. A word like "the" is therefore two times more frequent than the word "of" in standard English use. It is interesting to note that the frequency distribution of words in our corpus follows the Zipf law quite closely, including the very specific words to Twitter like 'rt' (for retweet).

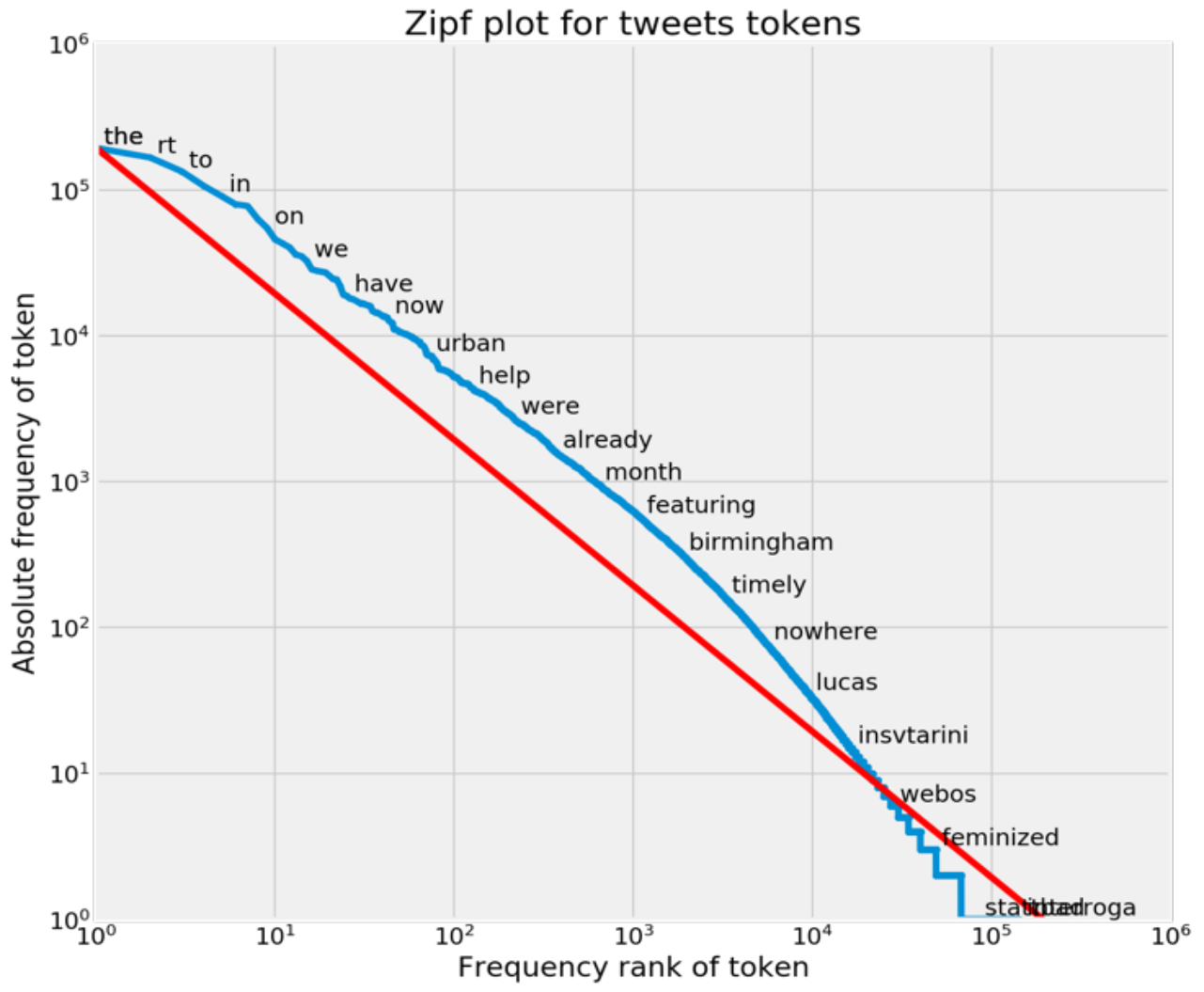

**Figure 2: Zipf's Law and frequency distribution of words in our corpus**

*4.1.2. Corpus Passport: Most Frequent Words*

The list of most frequent words identified in the corpus are clearly relevant to the state of today's media architecture practice: 'great', 'new', 'design' appears very frequently, as well as 'city', 'urban', 'people', 'public' and 'placemaking'. Frequency of words 'new' and 'future' indicate an orientation. Figure 3 illustrates the frequency of thirty most used words in the discourse.

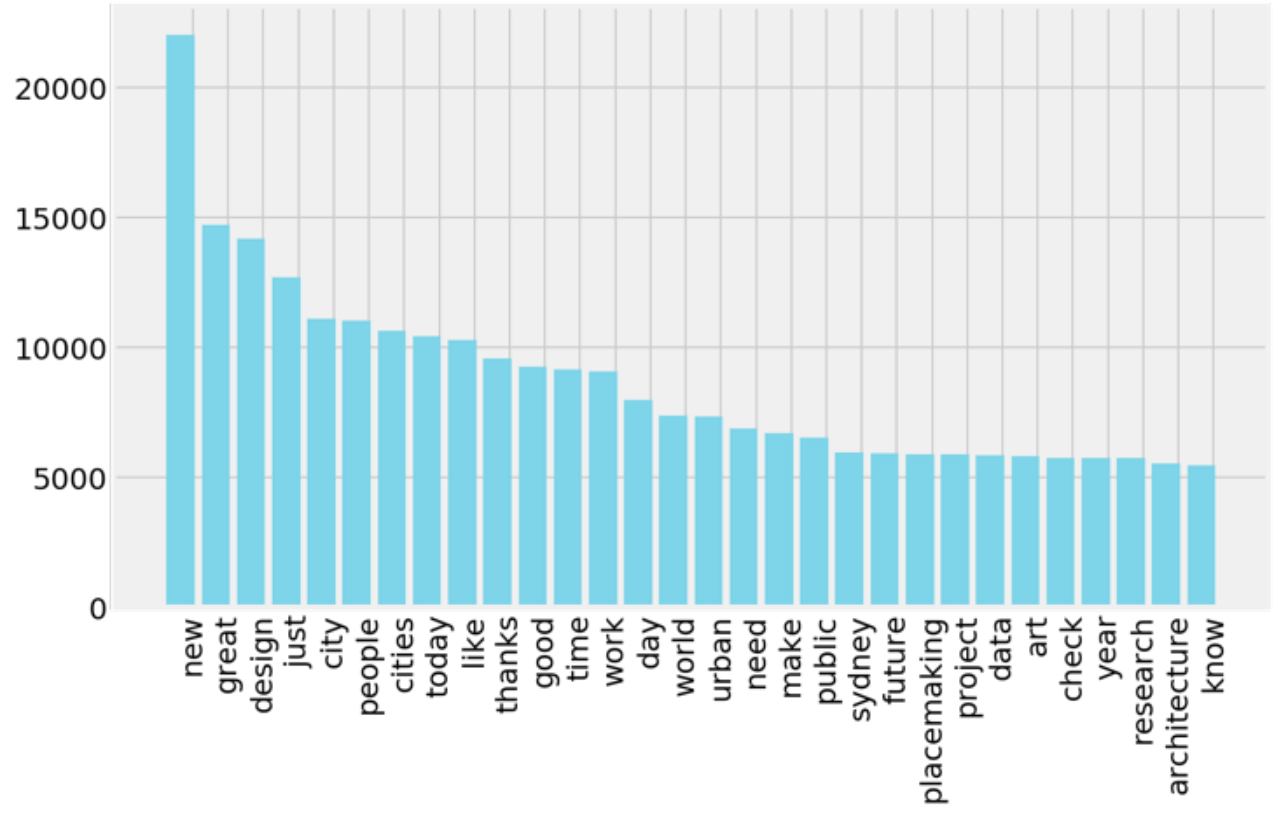

**Figure 3: The thirty most frequent words appearing in the corpus**

We looked at words that appear next to the 30 most frequent words. We gathered a list of co-occurrences which already suggests some of the emerging topics through grouping of short phrases. Topic modelling will be described later in the text, but it is interesting to observe whether and how they overlap.

[9] Zachary Booth Simpson's Vocabulary Analysis is available here : http://www.mine-control.com/zack/guttenberg/ (accessed May 4th 2018)

From the co-occurrences of the word ‘design’ we can observe that it is strongly associated with the web (and less e.g. design of physical interfaces, screens, facades, buildings).

['website designs', 'designs are', 'new website designs', 'website designs are', 'designs are almost', 'for design', 'design more', 'randomness for design', 'for design more', 'design more info']

Even more surprisingly, The word ‘city’ is associated with gender and ethnicity. Cities, on the other hand, are about sharing and activating.

['gender ethnicity', 'ethnicity related', 'to gender ethnicity', 'gender ethnicity related', 'ethnicity related privileges', 'nice city', 'city except', 'what nice city', 'nice city except', 'city except for']

['visible cities', 'cities prototyping', 'in visible cities', 'visible cities prototyping', 'cities prototyping lab', 'sharing cities', 'cities activating', 'of sharing cities', 'sharing cities activating', 'cities activating the']

## 4.2. Hypothesizing the question

In order to get a better sense of the relevance of these results compared to more general literature, and to check how our questions resonate with it, we compared the frequency of certain identified keywords in books scanned by Google Books project, using Google n-gram search[10]. Figure 4 shows the trend for the phrase ‘media architecture’ which, as can be expected, raises sharply since the middle of 1990s. The sharp decline we can observe since mid 2000s can perhaps be explained by the availability of data (most recent books may be less represented in the corpus), but it is certainly something to return to in the future.

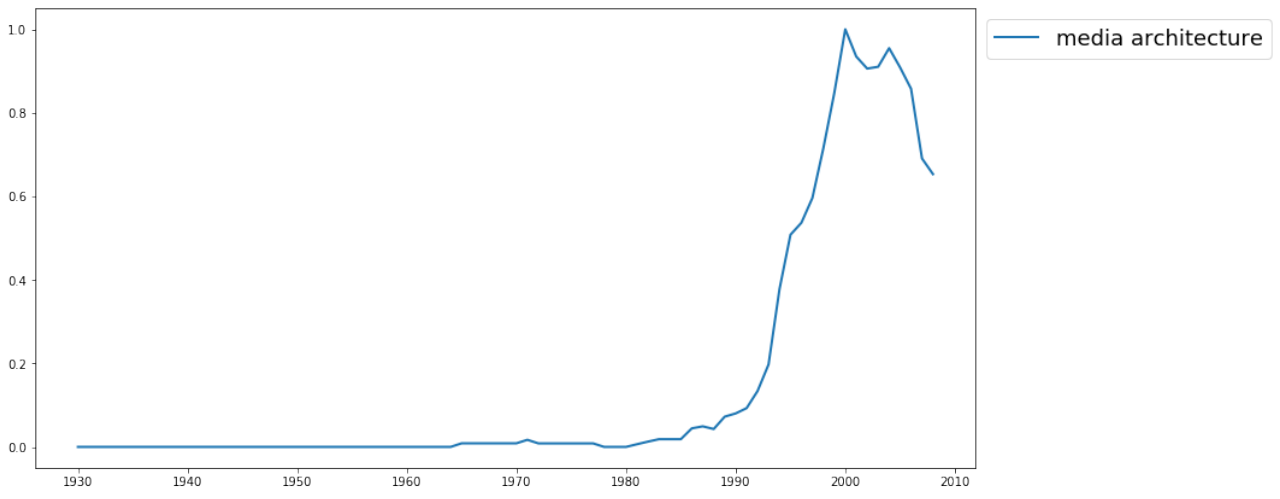

**Figure 4: Google n-gram trend for ‘media architecture’**

Furthermore, we look at frequency trends for some of the words from our most frequent list: 'architecture', 'design', 'city', 'urban', 'people', 'public', 'future', 'new' (Figure 5).

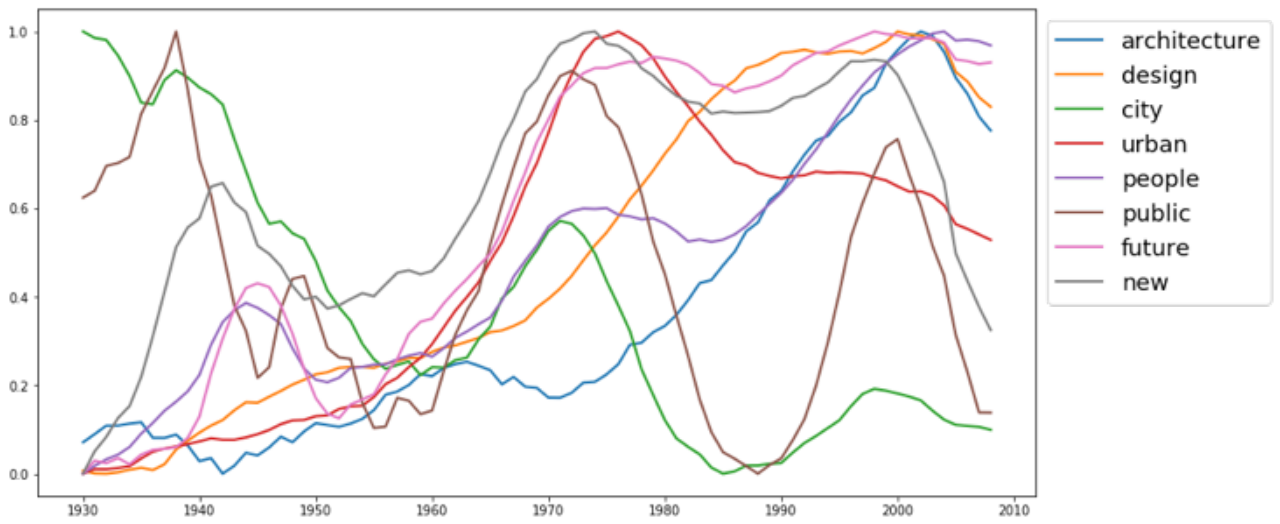

**Figure 5: Google n-gram trends since 1930 until today, for certain of the most frequent words from the corpus.**

If we observe these words in couples, we can notice some interesting interplay in their alternating popularity. ‘Architecture’ and ‘design’ have exchanged places in the end of 1930s and have only met at the same point in early 2000s. Since 1960s we talk (or at least write) more about the ‘urban’ then about the ‘city’. ‘People’ and ‘public’ oscillate in similar intervals, with ‘public’ reaching much more explicit peaks in the end of 1930s, early 1970s and 2000s. ‘Future’ and ‘new’ follow almost the same trend.

When compared to the more frequent terms in our corpus (Figure 3), the selected terms in Google Books project do not follow the same trends (Figure 5, observe the order of words according to the ends of the timelines, all the way to the right). For example, ‘people’, ‘future’ and ‘architecture’ are more prominent in general literature than in our corpus, although we could expect they would be important. To the contrary, ‘new’, ‘design’ and ‘city’ are higher ranked in our corpus than in the general literature.

## 4.3. Self-organizing Topics and Opinions

### *4.3.1. Corpus Passport: Vectorization*

With the *word2vec* model, a neural network trained on the English tweets only, we explored similarity between words and groups of words. The model enabled us to look at sets of words that are most similar (closest in terms of vector space) to a certain word. It is a more interesting way of looking at word relationships than simple co-occurrence.

We used TSNE tool (t-distributed Stochastic Neighbor Embedding)[11] to visualize the space of similarity between a word and terms that are closest to it in the model space (top 20 words). We can observe which words are strongly related to architecture: Venezia and Venice – for the Architecture Biennale; facades – for the focus on media facades in this discourse; contemporary – for the focus on future and new.

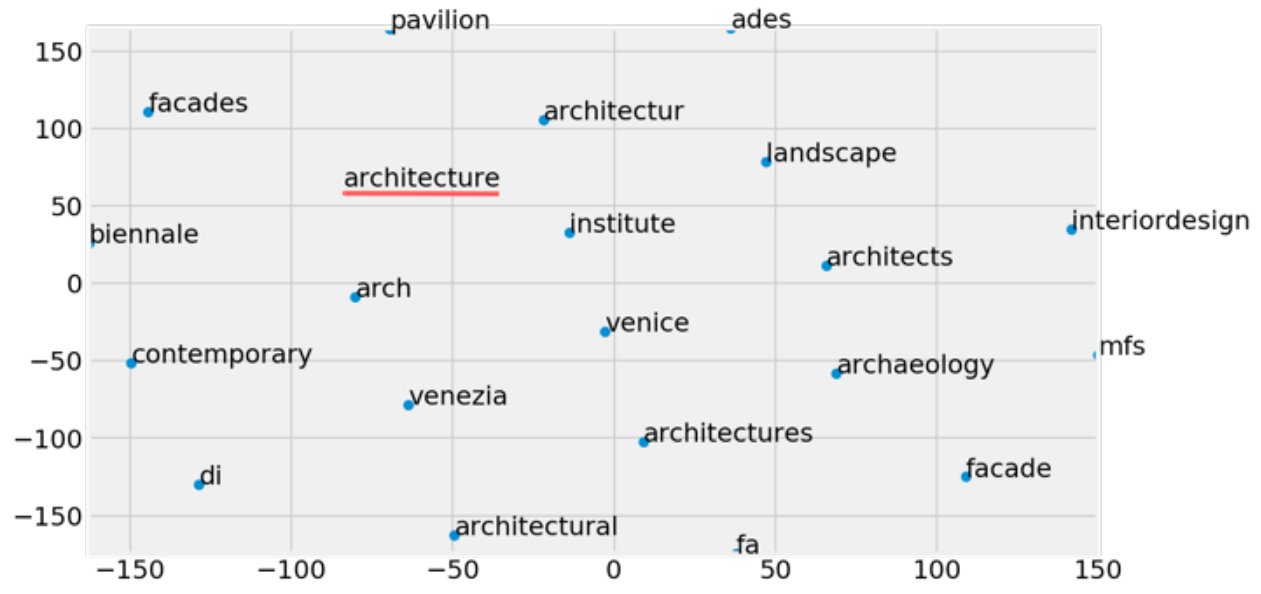

**Figure 6: Visualization of the space of similarity for the word ‘architecture’ throughout the corpus**

[10] Data on the frequency of different n-grams in books available within the Google's English text corpora

[11] t-SNE is a tool to visualize high-dimensional data. It converts similarities between data points to joint probabilities

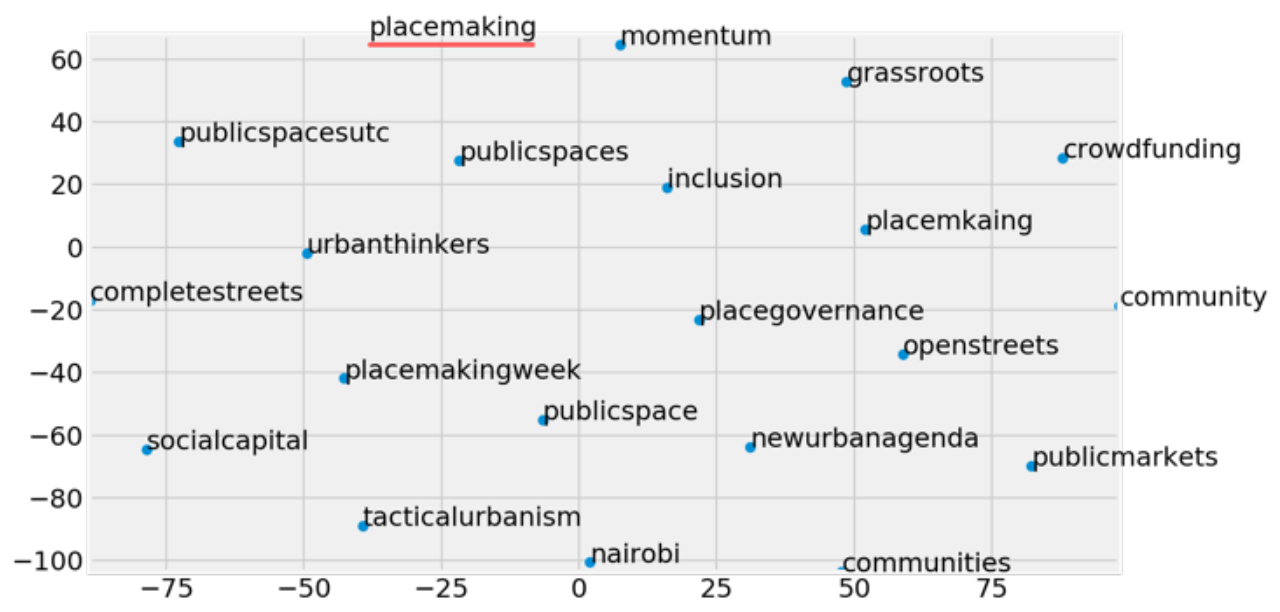

**Figure 7: Visualization of the space of similarity for the word "placemaking" throughout the corpus**

We also looked at other words that inform the discourse: 'design', 'city', 'public', 'infrastructure', 'future', 'placemaking'. On Figure 7 we can see

It is also possible to explore word similarity in contrast to some other words. So for example 'city' and 'placemaking' contrasted to 'space' gave us 'smartcity', while when contrasted to 'people' we got 'urbanism'. This is an interesting, open-ended way to explore the discourse, which demonstrates some of its implicit cultural preconceptions. It also sets ground for speculation about current trends that inform practice and policy of digital infrastructure making.

### 4.3.2. *Corpus Passport: Topic Modelling*

A different model is created to explore topic modelling of the discourse. We tested several standard models and algorithms that can generate them (within genism and scikit-learn Python modules). We had the best results with Latent Dirichlet Allocation Model from scikit-learn module, trained online. With this method, we identified ten most prominent topics, represented by keywords and their relative score in the model. These ten topics can be described as follows:

1. **Placemaking and education, research** ('research', 'project', 'great', 'night', 'placemaking', 'creative', 'workshop', 'home', 'like', 'talking', 'school')
2. **Arts, culture and digital technologies** ('people', 'art', 'best', 'digital', 'look', 'place', 'change', 'things', 'today', 'tech')
3. **Future urban living** ('cities', 'future', 'just', 'going', 'awesome', 'book', 'home', 'urban', 'life')
4. **Project management** ('project', 'check', 'week', 'want', 'know', 'good', 'year', 'let', 'free')
5. **Future health and trust** ('need', 'public', 'looking', 'join', 'make', 'smart', 'forward', 'health', 'today', 'game')
6. **Near future living – tech and place** ('just', 'space', 'did', 'big', 'right', 'got', 'talk', 'says', 'better', 'working')
7. **Architecture and innovation** ('new', 'architecture', 'love', 'read', 'world', 'social', 'media', 'building', 'innovation', 'students')
8. **Machine learning, AI and design** ('design', 'work', 'video', 'years', 'use', 'support', 'human', 'google', 'learning')
9. **Feelings for community** ('great', 'thanks', 'time', 'like', 'today', 'light', 'community', 'night', 'tonight', 'thank')
10. **Future city making** ('design', 'cities', 'urban', 'architecture', 'placemaking', 'future', 'big', 'building', 'years', 'better', 'lighting')

We then explored some of these topics in more detail, using an interactive graphical tool (Figure 8). The tool illustrates relative distances between identified topics, their representation in the discourse (size of the circle on the left) and keywords, with frequency, that describe it. For each of these topics, we can look at most prominent documents to better understand the context and content they represent.

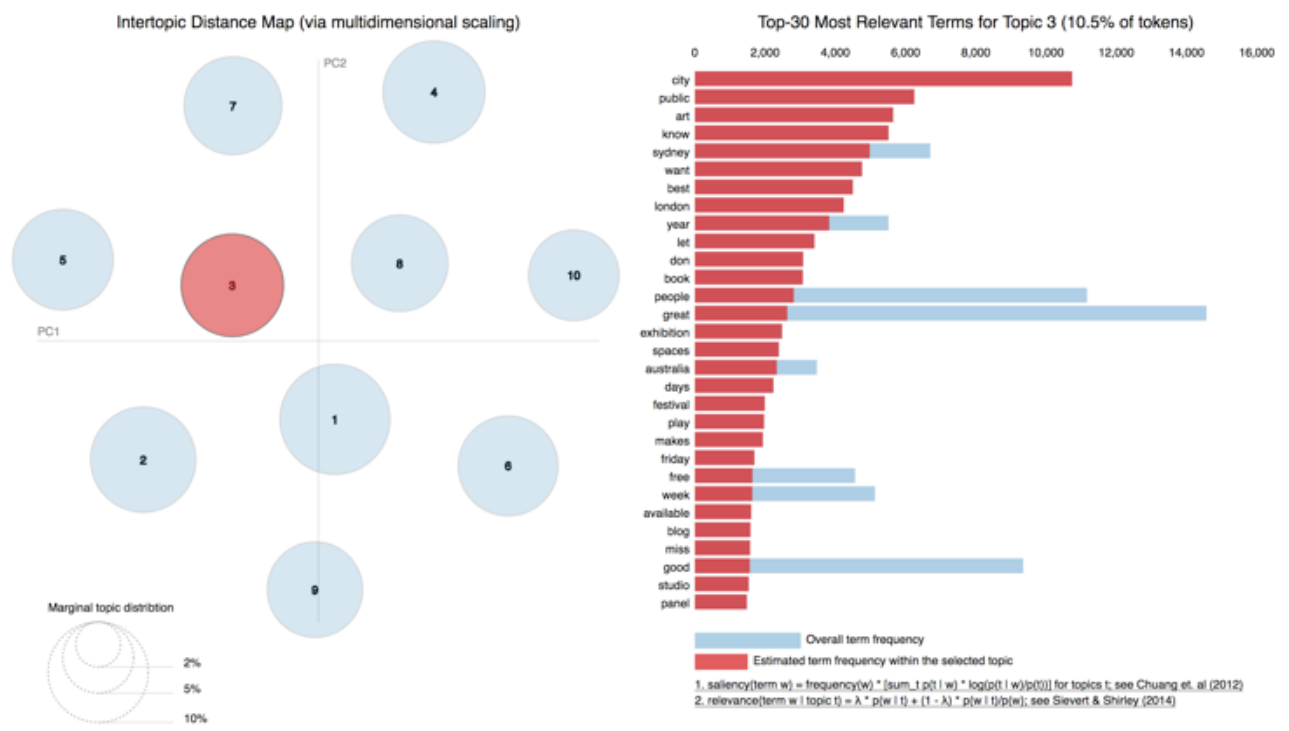

**Figure 8: Topic 3, 'Future urban living'. Position on the intertopic distance map and frequency of 30 most relevant terms**

### 4.3.1. *Looking at SOM Implementations*

We trained a SOM on the Word2vec and Doc2vec models, using the *sompy*[12] Python module.

We first look at the vector space of the corpus based on words through a U-matrix visualization (Figure 9). This U-matrix uses a divergent colour map, where blue hues can be thought of as clusters and yellow/red areas as cluster separators. In the yellow to red areas, there is a wider gap between the codebook values in the input space. Bright red circles are blobs of strong local similarity.

[12] Available at https://github.com/sevamoo/sompy

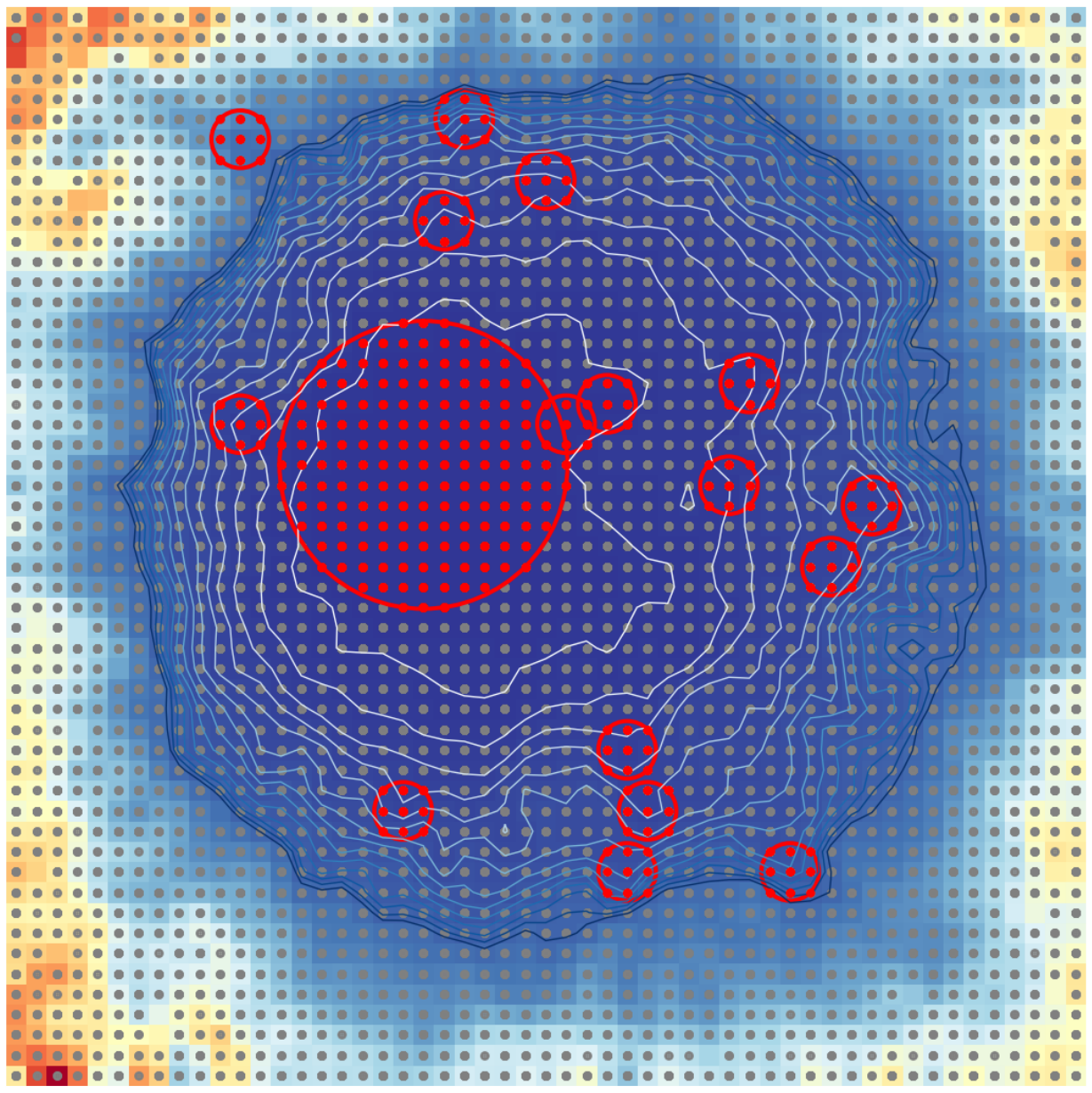

**Figure 9: SOM representation of word clusters trained on the word2vec model using *sompy***

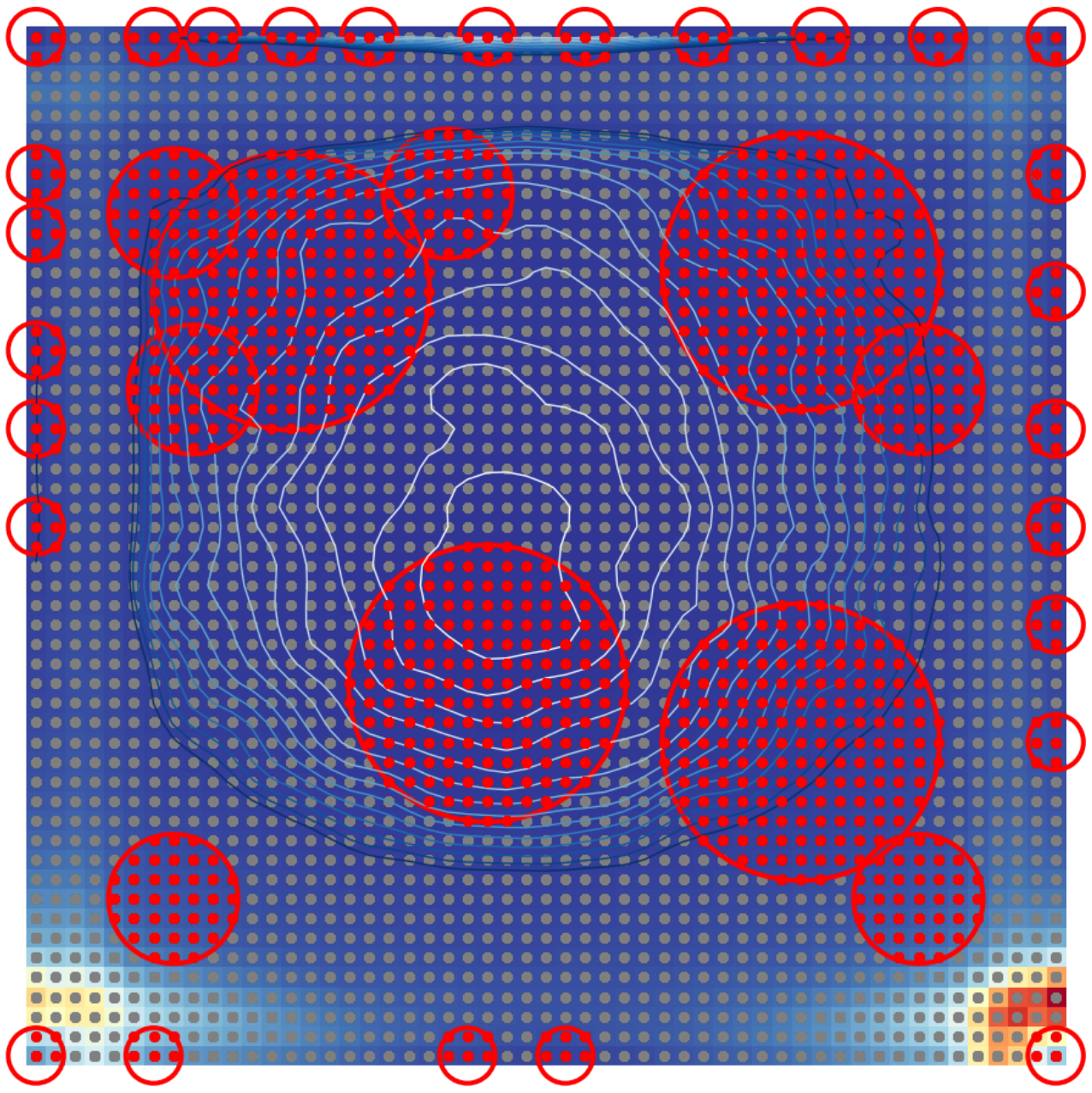

**Figure 10: SOM representation of document clusters trained on doc2vec model using sompy**

We then created a u-matrix using the SOM trained on documents (tweets). This visualization shows a very uniform situation, most of the space being clustered and gradually concentrating towards the centre.

We consider this to be very specific to the corpus examined here as this behaviour cannot be observed in other corpora we have examined (for example, a set of tweets from users that discuss digital tools and architecture).

## 5 CONCLUSION

The multitude of social media platforms creates unprecedent opportunities for communication and spontaneous exchange of opinions and experiences. It is typical for conferences and similar massively frequented events to be identified by a specific code, a *hashtag*, which helps aggregate this spontaneous exchange. We used this as an opportunity to identify a community of people who actively talk about media architecture on Twitter and explore this discourse. We gathered a large collection of posts and mined them for emerging keywords, topics and patterns.

Cutting away from the conventional perspective founded on ontologies of discourse fields, we look at the discourse about media architecture with the aim to identify self-organizing vectors of meanings. Our search for emergent topics and keywords, as well as patterns in discourse organisation (Figures 9 and 10) is attuned at identifying trends in the kinds of topics and the way they are discussed so that we could speculate on future trends and possible ways these findings could inform the practice.

The analysis presented here is not a typical humanities exploration of discourse attuned at revealing cultural values with a literary expertise. It is an attempt at working probabilistically with text, with the interest to create and instrument for inventive and projective work with this and other corpora. Our field of socialization and expertise is architecture and not social sciences, thus we were interested in the words that emerge from this discourse from the practice perspective.

In the statistical analysis of the discourse we looked at the discourse structure to identify significant characteristics. We found that the discourse is relatively rich in unique words, when compared to literature. We also found that the use of platform specific codes (hashtags, mentions) was relatively high.

The three most frequent words, ‘great’, ‘new’, ‘design’ do describe well the overall optimistic character of the corpus. Without going into the debate as to why or what the community is optimistic about, we can observe that a forward looking orientation towards new technologies and infrastructures is reflected in these and other frequently used terms (see Figure 3).

We were able to confirm these assumptions through topic modelling, where we identified ten most prominent emerging topics. *Placemaking and research*, *Near-future living* and *Future city making* illustrate this spirit well. Placemaking emerges as one of the most prominent words, and is present in two topics. Care or support for communities (social structures) are also prominent concepts. The fact that this reflects very well the theme of the upcoming Media Architecture Biennale should not be taken lightly. It is important to remind that our corpus is not made only of tweets that mention hashtags that we search for (which would lead this inquiry into a kind of self-fulfilling prophecy) but rather consists of entire bodies of tweets of all users that *mentioned* or simply *liked* tweets with the specific hashtag. Thus the tweets that contain the hashtag account for a very small portion of all tweets of a user (many having over 3000 tweets).

Finally, we looked at patterns that emerge from the corpus, using self-organizing map machine learning algorithm that maps similar words or documents closer together, creating clusters in this way. What we found striking in the maps is the uniformity of this organisation that characterizes the corpus, especially on the level of documents. We identify this to be a sign of discourse homogeneity, something that the community should seek to overcome in the future.

The approach attuned at emerging concepts and topics is very important for this work. Without preconceptions of categories and models our findings could fit to, we are free to ask different questions and have different thoughts – we can approach things differently.

This study offers a critical perspective on media architecture through the lens of how it has been talked about. We explored social and cultural aspects of the media architecture community’s discourse using text-mining and machine learning techniques. An exploratory approach informed by statistical analysis and topic modelling (LDA and SOM) was applied to social media conversation within the community interested in media architecture, to identify ways they talk about it, the trends, the topics, the concerns. We hope to have hinted at future trends or concerns by way of inventing an abstract way of looking at current ones.

In our future work, we envisage inventing ways to project out of this vector space of discourse by exposing it to external attractors. These external attractors would be informed by similar interests in technology and connectedness but from the perspective of mathematics and information theory. We expect that this would open up interesting ways out of homogeneity and hermeticity of discourse.

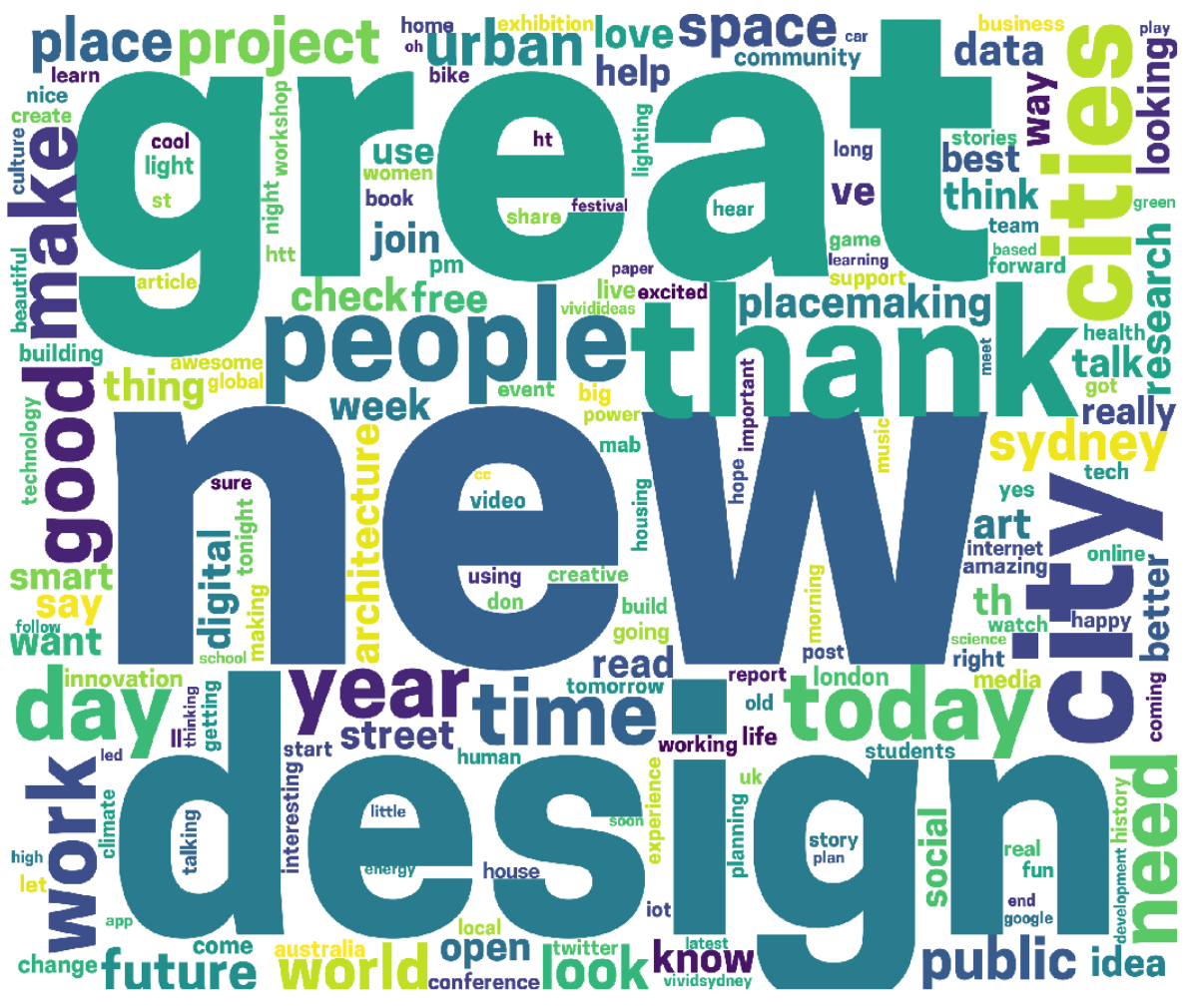

**Figure 11: An illustration of the prominence of 'great', 'new', 'design' in the corpus. Wordcloud of 200 most frequent terms.**

The contribution of this work to the Media Architecture community is in addressing its’ own basis – the way it is being discussed by the interested and invested community or practitioners and researchers. We hint at possible future trends and questions that will have to be addressed, on the line of topics we have identified, but informed by external inputs. We identify certain social and cultural aspects of the community through this analysis – the interests, the habits of expression, the homogeneity. We believe the interest in digital infrastructure and hybrid city can be equally well addressed from the perspective of how it is talked about as from the perspective of practical applications, and the former can enrich future discussions and point to new ideas.

## ACKNOWLEDGMENTS